\documentclass{article}

\PassOptionsToPackage{numbers, compress}{natbib}

\usepackage[preprint]{neurips_2024}

\usepackage[utf8]{inputenc} 
\usepackage[T1]{fontenc}    
\usepackage{hyperref}       
\usepackage{url}            
\usepackage{booktabs}       
\usepackage{amsfonts}       
\usepackage{nicefrac}       
\usepackage{microtype}      
\usepackage{xcolor}         
\usepackage{bm}
\usepackage{amsmath}
\usepackage{algorithm}
\usepackage{algorithmic}
\usepackage{graphicx}
\usepackage{multirow}
\usepackage{subfig}

\title{Hashing based Contrastive Learning \\for  Virtual Screening}

\author{%
Jin~Han\thanks{Equal contribution}, Yun~Hong\footnotemark[1], Wu-Jun~Li\thanks{Corresponding author} \\
National Key Laboratory for Novel Software Technology\\
  Department of Computer Science and Technology\\
  Nanjing University, Nanjing 210023, China\\
  \texttt{\{hanjin, hongy\}@smail.nju.edu.cn,liwujun@nju.edu.cn} \\
}

\begin{document}

\maketitle

\begin{abstract}
Virtual screening (VS) is a critical step in computer-aided drug discovery, aiming to identify molecules that bind to a specific target receptor like protein. 
Traditional VS methods, such as docking, are often too time-consuming for screening large-scale molecular databases. 
Recent advances in deep learning have demonstrated that learning vector representations for both proteins and molecules using contrastive learning can outperform traditional docking methods.
However, given that target databases often contain billions of molecules, real-valued vector representations adopted by existing methods can still incur significant memory and time costs in VS. 
To address this problem, in this paper we propose a hashing-based contrastive learning method, called DrugHash, for VS.
DrugHash treats VS as a retrieval task that uses efficient binary hash codes for retrieval. In particular, DrugHash designs a simple yet effective hashing strategy to enable end-to-end learning of binary hash codes for both protein and molecule modalities, which can dramatically reduce the memory and time costs with higher accuracy compared with existing methods.
Experimental results show that DrugHash can outperform existing methods to achieve state-of-the-art accuracy, with a memory saving of 32$\times$ and a speed improvement of 3.5$\times$.
\end{abstract}

\section{Introduction}
\label{introduction}
Due to the high failure rates of drug candidates, drug discovery requires a long development cycle and substantial costs~\cite{gorgulla2020open}.
Virtual screening (VS) is a critical step in computer-aided drug discovery, aiming to identify molecules that bind to a specific target receptor.
Representative receptors include biological macromolecules such as protein, DNA and RNA.
High-quality VS could identify promising lead compounds, thereby reducing time and resource costs in drug discovery. 
The size of the molecular database is an important factor affecting the effectiveness of VS. 
Increasing the database size can typically include more candidate molecules. However, as shown in~\cite{lyu2019ultra}, a poorly performing VS method will increase the false positive rate when the database is enlarged. This poses two requirements for effective VS: a large-scale molecular database and an effective VS algorithm.

Numerous large-scale molecular databases have been developed for drug discovery.
For instance, ZINC~\cite{irwin2005zinc}, a commercially available molecular database, boasts a vast collection of over 1.4 billion compounds in its latest version ZINC20~\cite{irwin2020zinc20}. The Enamine REAL database~\cite{shivanyuk2007enamine} serves as a robust tool for large-scale virtual screening, featuring a substantial repository of over 6.75 billion molecules in its current release. However, despite the availability of these diverse databases, existing VS methods are not equipped to screen such large-scale databases. Traditional VS methods, such as molecular docking, are often too time-consuming for screening large-scale molecular databases. 
Taking the AutoDock Vina~\cite{kitchen2004docking} as an example, when the exhaustiveness is set to 8, it takes about 70 seconds to dock a molecule on a single-core CPU.
Hence, docking millions of molecules, the scale of which is comparable with or even smaller than those in many real applications, would take Vina more than two years. 
Although increasing the number of CPUs can reduce docking time, this inevitably leads to substantial resource consumption.
The learning-based VS methods~\cite{zheng2019onionnet,jones2021improved,zhang2023planet,wu2022bridgedpi,yazdani2022attentionsitedti} are relatively resource-effective.
However, most of these methods attempt to predict the binding affinity or interactions between the protein and molecule, which fail to outperform docking methods on VS benchmarks. 

Recently, DrugCLIP~\cite{gao2024drugclip} proposes to treat VS as a retrieval task rather than learning the binding affinity or interactions of protein-molecule pairs. DrugCLIP allows for the use of more unlabeled training data and enables the pre-encoding of molecular representations into vectors to accelerate the retrieval process. By learning real-valued vector representations for both proteins and molecules through contrastive learning, DrugCLIP can outperform traditional docking methods. However, given that target databases often contain billions of molecules, the real-valued vector representations for molecular databases can still incur significant memory costs in VS. For example, DrugCLIP encodes molecules into 128-dimensional real-valued vectors, and hence the real-valued vector representations for the Enamine REAL database would have a memory cost of more than 3TB, which is extremely large for local computer memory. Moreover, calculating the similarity between billions of real-valued vectors and ranking the results is time-consuming.

In this paper, we propose a hashing-based contrastive learning method, called DrugHash, for VS.
DrugHash also treats VS as a retrieval task, but it uses efficient binary hash codes for retrieval.
The contributions of DrugHash are outlined as follows:
\begin{itemize} 
\item To the best of our knowledge, DrugHash is the first hashing method for VS. 
\item DrugHash designs a simple yet effective hashing strategy to enable end-to-end learning of binary hash codes for both protein and molecule modalities, which can dramatically reduce the memory and time costs.
\item  DrugHash can also outperform existing methods in terms of accuracy. This seems to be counter-intuitive but is actually reasonable, because binary hash codes can act as a constraint~(regularization) for improving generalization ability.
\item Experimental results show that DrugHash can outperform existing methods to achieve state-of-the-art accuracy, with a memory saving of 32$\times$ and a speed improvement of 3.5$\times$.
\end{itemize}

\section{Related Works}
\paragraph{Virtual Screening}
Existing VS methods can be categorized into two main classes: docking-based methods and learning-based methods. Docking-based methods like FlexX~\cite{rarey1996fast}, Glide~\cite{friesner2004glide} and AutoDock Vina~\cite{trott2010autodock} involves predicting the molecule conformation and pose given a target protein~\cite{kitchen2004docking}. Docking relies on complex scoring functions to estimate the binding affinity or strength of the connection across the molecule and the target protein~\cite{raval2022basics}. Random approaches like Monte Carlo and genetic algorithms are performed to search the vast conformation space to find the optimal docking pose. Hence, docking-based methods are typically very time-consuming. 

Learning-based methods can be further categorized into supervised and unsupervised methods. Supervised methods are supervised by the binding affinities or interactions between molecules and proteins. Methods such as OnionNet~\cite{zheng2019onionnet}, FAST~\cite{jones2021improved}, Planet~\cite{zhang2023planet}, and CAPLA~\cite{jin2023capla} treat the problem as a regression task that predicts the binding affinities of the protein-ligand complex. On the other hand, a bunch of methods such as BridgeDPI~\cite{wu2022bridgedpi}, TransformerCPI~\cite{chen2020transformercpi}, and AttentionSiteDTI~\cite{yazdani2022attentionsitedti} predict whether or not a molecule could bind to the target protein, which is a binary classification task. However, due to the shortage of labeled data, the performance of these supervised methods is limited~\cite{brocidiacono2022bigbind}. Recently, DrugCLIP~\cite{gao2024drugclip} proposes to treat VS as a retrieval task which is unsupervised. By learning real-valued vector representations for proteins and molecules in a contrastive learning approach and applying data argumentation, DrugCLIP can surpass traditional docking methods on VS benchmarks. However, the real-valued vectors for feature representations will bring large memory and time costs to screen billion-scale molecular databases.

\paragraph{Cross-Modal Hashing}
Hashing adopts a hash function to map each data point in the database to a binary vector~(or called hash code) while preserving the similarity in the original space. In cross-modal hashing~(CMH), the modality of a query point is different from the modality of the points in the database, and hence different hash functions will be designed for different modalities~\cite{bronstein2010data, kumar2011learning, lin2015semantics}. Binary hash codes have advantages in low storage cost and fast retrieval time, and hence there are many CMH methods proposed for multimedia data~(text, audio, images, video) retrieval~\cite{jiang2017deep, deng2018triplet, shen2020exploiting, hu2022unsupervised}. To the best of our knowledge, no research has explored hashing and CMH on VS problems including protein-molecule retrieval. 

\section{Method}
\label{method}
\begin{figure}[t]
\centering
\includegraphics[scale=0.57]{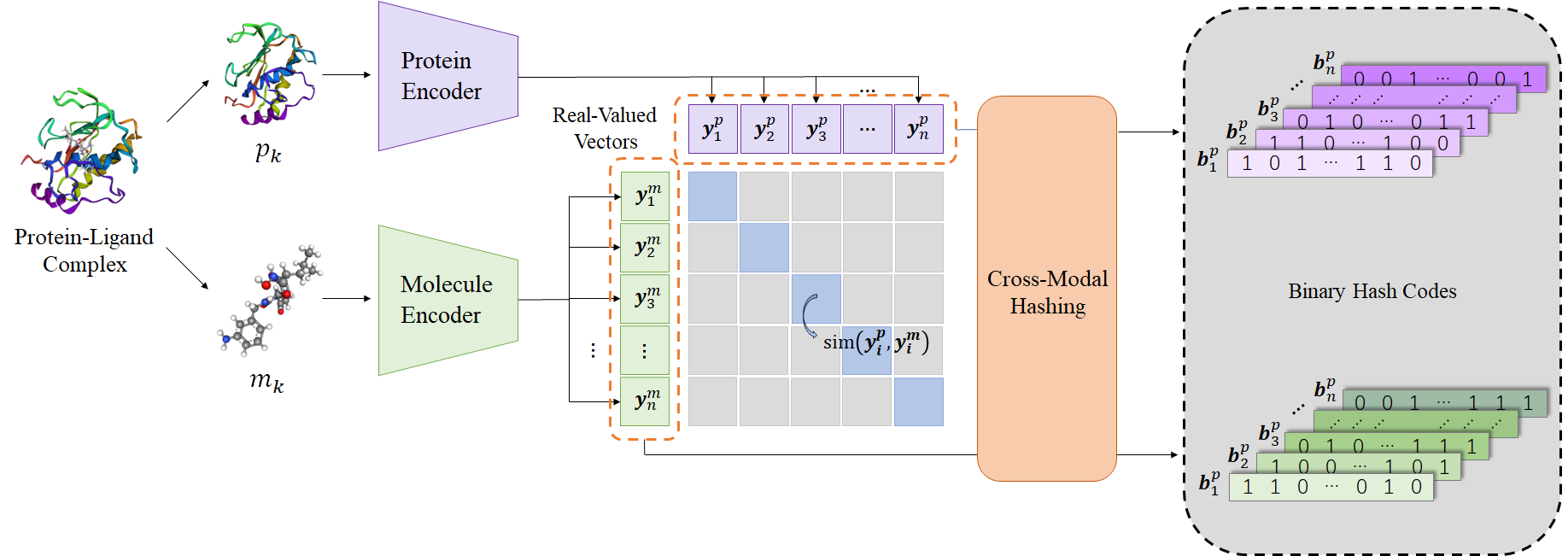}
\caption{Illustration of the DrugHash architecture.}
\label{figure1}
\end{figure}

In this section, we present the details of our proposed method called DrugHash, which is a hashing-based contrastive learning method.
DrugHash treats VS as a retrieval task. Although the techniques in DrugHash can be used for various VS problems, we focus on protein-ligand complexes~\cite{wang2005pdbbind,yang2012biolip,gaulton2012chembl} in this paper. Hence, the term protein actually refers to the protein pocket. For brevity, we will continue to use protein in the following content.
The term ligand refers to a small molecule that binds to a target protein.
The complexes reveal the structural and functional relationships between proteins and molecules. 
We treat the complexes as data of two modalities: proteins and molecules. DrugHash uses protein queries to retrieve molecules in the database.

The hash function can be manually designed or learned from the training data. We focus on learning hash functions for proteins and molecules. Figure~\ref{figure1} illustrates the architecture of DrugHash\footnote{The protein and molecule pictures in Figure~\ref{figure1} and Figure~\ref{figure2} are from Protein Data Bank~\cite{berman2000protein}.}. DrugHash contains the protein and molecule encoder and the objective function which includes a contrastive learning objective and a cross-modal hashing objective.


\subsection{Protein and Molecule Encoder}
Various protein and molecule encoders can be adopted in DrugHash. To show the effectiveness of our proposed hashing strategy, we adopt the same encoder as DrugCLIP~\cite{gao2024drugclip}, which is a pre-trained SE(3) Transformer proposed in Uni-Mol~\cite{zhou2023uni}. The input of the encoder are atom representation which is initialized from atom type and pair representation which is initialized by encoding the Euclidean distance between pairs of atoms using a Gaussian kernel~\cite{shuaibi2021rotation}. We denote the pair representation of $ij$ in layer $l$ as $\bm{q}_{ij}^l$. $\bm{q}_{ij}^l$ is first updated by the Query-Key product in attention mechanism~\cite{vaswani2017attention}:
\begin{equation}
    \begin{aligned}
    \bm{q}_{ij}^{l+1}=\bm{q}_{ij}^{l}+\{\frac{\bm{Q}_i^{l,h}(\bm{K}_j^{l,h})^\top}{\sqrt{d}}\mid h\in\lbrack 1, H\rbrack\},\\
    \end{aligned}
\end{equation}
where $\bm{Q}_i^{l,h}$ and $\bm{K}_j^{l,h}$ is the Query and Key of $i$-th and $j$-th atom representation in layer $l$ for the $h$-th attention head, $H$ is number of attention heads, $d$ is the dimension of Key. The pair representation serves as a bias term in self-attention to update the atom representation:
\begin{equation}
    \mathrm{Attention}(\bm{Q}_i^{l,h}, \bm{K}_j^{l,h}, \bm{V}_j^{l,h})=\mathrm{softmax}(\frac{\bm{Q}_i^{l,h}(\bm{K}_j^{l,h})^\top}{\sqrt{d}}+\bm{q}_{ij}^{l-1,h})\bm{V}_j^{l,h},
\end{equation}
where $\bm{V}_j^{l,h}$ is the Value of $j$-th atom representation in layer $l$ for the $h$-th attention head.  The above encoder is pre-trained through 3D position recovery and atom-type recovery tasks. 
The input atom coordinates are randomly corrupted, and the model is trained to predict the correct positions and pairwise distance of the atoms.
The atom types of corrupted atoms are also masked by a $[\mathrm{CLS}]$ token, and the model is trained to predict the masked atom type.
The protein encoder is pre-trained on 3.2 million protein pockets and the molecule encoder is pre-trained on 19 million molecules. The encoder of DrugHash is initialized by the parameter of the pre-trained encoder.

Formally, we denote the above encoding process of protein and molecule as $E_p$ and $E_m$ respectively. Considering a set of $n$ complexes, we  denote the set of proteins as $P=\{p_1, p_2, \cdots, p_n\}$ and the set of molecules as $M=\{m_1, m_2, \cdots, m_n\}$.
For any index $k$, $p_k$ and $m_k$ indicate the protein and molecule data originating from the same complex. For a protein-molecule pair $(p_k, m_k)$, their vector representation $(\bm{y}_k^p,\bm{y}_k^m)$ could be obtained by:
\begin{equation}
    (\bm{y}_k^p,\bm{y}_k^m)=(E_p(p_k), E_m(m_k)).
\end{equation}
\subsection{Objective Function}
Our objective function includes a contrastive learning objective and a cross-modal hashing strategy. The contrastive learning objective aims to align the representation of proteins and molecules in a shared embedding space. The cross-modal hashing strategy aims to enable end-to-end learning of the binary hash codes for both protein and molecule modalities.
\paragraph{Contrastive Learning}
 We aim to maximize the similarity between correct pairs and minimize the similarity between incorrect pairs. In this context, the word ``correct'' means that the protein and molecule belong to the same complex. The encoded protein set $E_p(P)=\{\bm{y}_1^p, \bm{y}_2^p, \cdots, \bm{y}_n^p\}$ and the encoded molecule set $E_m(M)=\{\bm{y}_1^m, \bm{y}_2^m, \cdots, \bm{y}_n^m\}$. As shown in Figure~\ref{figure1}, there are $n^2$ protein-molecule pairs in total, but only the pairs in the diagonal are supposed to be similar.
We use cosine similarity to define the similarity between $\bm{y}_i^p$ and $\bm{y}_j^m$:
\begin{equation}
    \mathrm{sim}(\bm{y}_i^p,\bm{y}_j^m)=\bm{y}_i^p\cdot (\bm{y}_j^m)^\top/(\Vert \bm{y}_i^p\Vert \Vert \bm{y}_j^m\Vert).
\end{equation}
 We use infoNCE loss \cite{oord2018representation} to define our contrastive learning objective, which aims to minimize the negative log-likelihood of similar protein-molecule pairs. From the perspective of protein modality, we aim to distinguish the true ligand that binds to the given protein:
\begin{equation}
    \begin{aligned}
        \mathcal{L}_k^p&=-\frac{1}{n}\log\frac{\exp(\mathrm{sim}(\bm{y}_k^p, \bm{y}_k^m)/\tau)}{\sum_i\exp(\mathrm{sim}(\bm{y}_k^p, \bm{y}_i^m)/\tau)},\\
    \end{aligned}
\end{equation}
where $\tau$ denotes a temperature hyperparameter. From the perspective of molecule modality, we aim to distinguish the true receptor that accepts the given molecule:
\begin{equation}
    \begin{aligned}        
        \mathcal{L}_k^m&=-\frac{1}{n}\log\frac{\exp(\mathrm{sim}(\bm{y}_k^p, \bm{y}_k^m)/\tau)}{\sum_i\exp(\mathrm{sim}(\bm{y}_i^p, \bm{y}_k^m)/\tau)}.\\        
    \end{aligned}
\end{equation}
The above loss function has been utilized in previous works like CLIP~\cite{radford2021learning} and DrugCLIP~\cite{gao2024drugclip}. The overall contrastive learning objective is defined as:
\begin{equation}
    \begin{aligned}
        \mathcal{L}_c &= \frac{1}{2}\sum_{k=1}^n(\mathcal{L}_k^p+\mathcal{L}_k^m).\\
    \end{aligned}
\end{equation}

\paragraph{Cross-Modal Hashing}
\label{p1}
Note that the vector representation $\bm{y}_k^p$ and $\bm{y}_k^m$ of protein and molecule is still real-valued at this stage. Unlike the previous methods,  we aim to learn the binary hash codes for both protein and molecule modality. Let $\bm{b}_k^p\in \{-1, 1\}^d$ and $\bm{b}_k^m\in\{-1,1\}^d$ denote the binary hash codes for protein $p_k$ and molecule $m_k$, where $d$ is the code length, which is the same as the embedding dimension of $\bm{y}_k$. We define our loss function as follows:
\begin{equation}
    \begin{aligned}
        \mathcal{L}_{\mathrm{hash}} &= \frac{1}{nd}\sum_{k=1}^n(\Vert \bm{y}_k^p-\bm{b}_k^p\Vert_2^2+\Vert\bm{y}_k^m-\bm{b}_k^m\Vert_2^2),\\        
    \end{aligned}
\end{equation}
 The loss function designed in this way has two purposes. On one hand, it brings $\bm{y}_k^p$ and $\bm{y}_k^m$ closer to its binary hash codes $\bm{b}_k^p$ and $\bm{b}_k^m$. On the other hand, it serves as a regularization term to reduce model overfitting.
 
 The overall objective function is formulated as follows:
\begin{equation}
    \begin{aligned}
        \mathcal{L}&=\mathcal{L}_c+\lambda \mathcal{L}_{\mathrm{hash}},\\
    \end{aligned}
\end{equation}
where $\lambda$ is a hyperparameter for balancing the two loss items.
\subsection{Training and Inference}
We denote the whole parameter of $E_p$ and $E_m$ as $\theta$. In the training stage, the model parameter $\theta$ and binary hash codes $\bm{b}_k^p$ and $\bm{b}_k^m$ can be optimized alternately.
When $\theta$ is fixed, the $\bm{b}_k^p$ and $\bm{b}_k^m$ could be obtained by 
\begin{equation}
    \begin{aligned}
        \bm{b}_k^p &= \mathrm{sign}(\bm{y}_k^p),\\
        \bm{b}_k^m &= \mathrm{sign}(\bm{y}_k^m),\\
    \end{aligned}
\end{equation}
where the function $\mathrm{sign}(\cdot)$ returns the signs of the elements. When the binary hash codes are fixed, $\theta$ can be optimized by backpropagation.

\begin{figure}[t]
\centering
\includegraphics[scale=0.35]{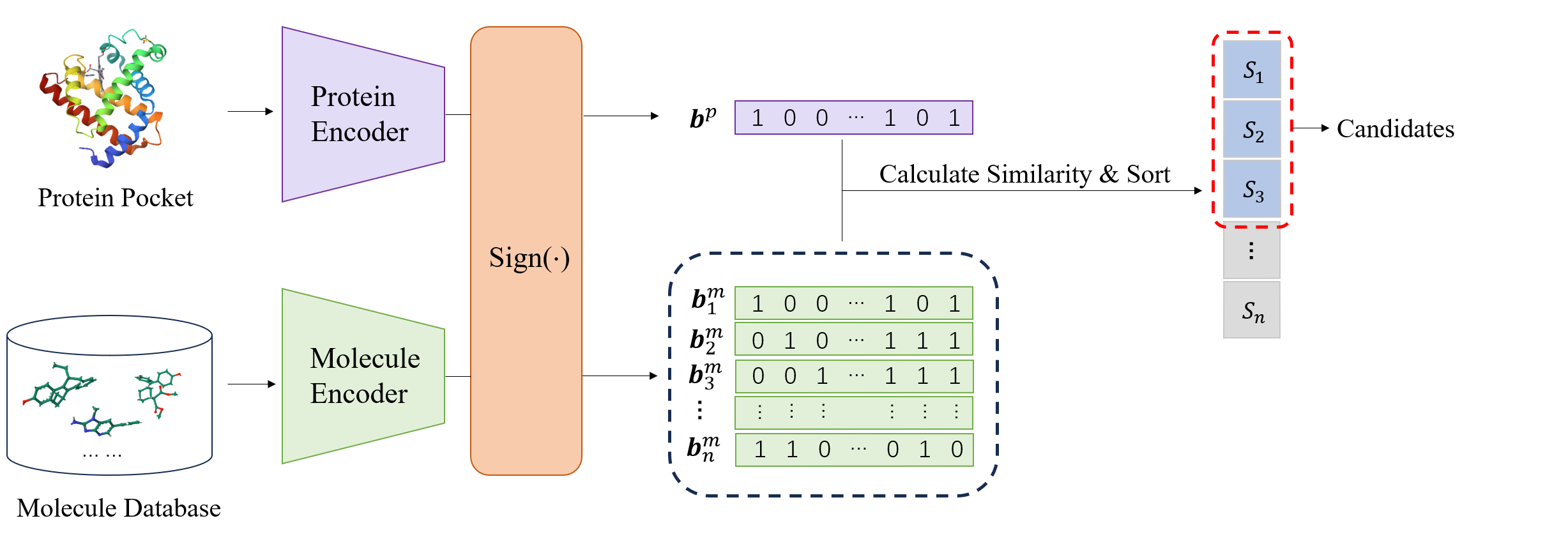}
\caption{Illustration of the inference phase.}
\label{figure2}
\end{figure}

In the inference stage, as illustrated in Figure~\ref{figure2}, we aim to retrieve molecules that would bind to the target protein from the molecular database. Given a target protein $p$ and molecular database $D_M=\{m_1, m_2,m_3,\cdots\}$, the binary hash code $\bm{b}^p$ of protein and binary vector database $B_{D_m}$ of molecules can be obtained as: 
\begin{equation}
    \begin{aligned}
        \bm{b}^p   &=\mathrm{sign}(E_p(p)),\\
        B_{D_m}    &=\{\bm{b}_1^m, \bm{b}_2^m, \bm{b}_3^m, \cdots\}
                   =\{\mathrm{sign}(E_m(m_1)), \mathrm{sign}(E_m(m_2)), \mathrm{sign}(E_m(m_3)), \cdots\}.\\
    \end{aligned}
\end{equation}
The number -1 in hash codes can be easily changed to 0 to get the final hash code representation.
We rank the molecules most likely to bind with the target protein based on Hamming distance, which can be calculated as the different bits of binary hash codes.

\section{Experiment}
\label{experiment}
\subsection{Evaluation Settings}
\paragraph{Metric}
We adopt several metrics to evaluate the accuracy of VS: the area under the receiver operating characteristic curve (AUROC), the Boltzmann-enhanced discrimination of receiver operating characteristic (BEDROC), and the enrichment factor (EF). AUROC is a commonly used metric to evaluate the ranking performance. BEDORC is a metric proposed in \cite{truchon2007evaluating}, which is sensitive to "early recognition". The BEDROC is formulated as $\mathrm{BEDROC}=\frac{\sum_{i=1}^ne^{-\alpha r_i/N}}{R_\alpha(1-e^{-\alpha})/(e^{\alpha/N-1})}\times \frac{R_\alpha sinh(\alpha/2)}{cosh(\alpha/2)-cosh(\alpha/2-\alpha R_\alpha)}+\frac{1}{1-e^{\alpha (1-R_\alpha)}}$, where $n$ is the number of actives, $N$ is the total number of molecules, $R_\alpha=n/N$, $r_i$ is the rank position of the 
i-th active and $\alpha$ is set to 80.5 in our experiment. EF is formulated as $\mathrm{EF}^{x\%}=\frac{\mathrm{Hits}^{x\%}/\mathrm{N}^{x\%}}{\mathrm{Hits}^{total}/\mathrm{N}^{total}}$, where $\mathrm{Hits}^{x\%}$ and $\mathrm{Hits}^{total}$ means the actives in top $x\%$ and total database, $\mathrm{N}^{x\%}$ and $\mathrm{N}^{total}$ means the number of molecules in top $x\%$ and total database. We consider the $\mathrm{EF}^{x\%}$ of $\mathrm{EF}^{0.5\mathrm{\%}}$, $\mathrm{EF}^{1\%}$ and $\mathrm{EF}^{5\%}$. We also calculate the memory and time cost for different VS methods.
\paragraph{Datasets} 
To train DrugHash, we employ the same training data as DrugCLIP, which is the PDBBind~\cite{wang2005pdbbind} dataset argumented by HomoAug~\cite{gao2024drugclip}. 
To benchmark the VS performance of different methods, we employ two evaluation datasets, which are DUD\--E~\cite{mysinger2012directory} and LIT\--PCBA~\cite{tran2020lit}. DUD\--E contains 102 protein targets, each target is associated with a set of ligands and decoys. The total number of ligands is 22,886 each with 50 property-matched decoys. LIT\--PCBA contains 15 target proteins, and the total number of active compounds is 7,844 while that of inactive compounds is 407,381. To evaluate the memory and time cost of different VS methods, we employ the ZINC~\cite{irwin2020zinc20} database and the REAL database of Enamine~\cite{shivanyuk2007enamine}. Zinc contains 2.3 million ready-to-dock molecules. Enamine REAL is a database designed for ultra-large scale VS which contains more than 6.75B molecules.
\paragraph{Baselines}
On the DUD\--E benchmark, we adopt 7 baselines, which include two docking methods Glide\--SP~\cite{friesner2004glide}and Vina~\cite{trott2010autodock}, and five learning based methods NN-score~\cite{durrant2011nnscore}, RF\--Score~\cite{ballester2010machine}, Pafnucy~\cite{stepniewska2017pafnucy}, Planet~\cite{zhang2023planet},Banana~\cite{brocidiacono2022bigbind}, and DrugCLIP~\cite{gao2024drugclip}. On the LIT\--PCBA benchmark, we adopt 7 baselines, which include three docking methods Surflex~\cite{spitzer2012surflex}, Glide\--SP~\cite{friesner2004glide} and Gnina~\cite{mcnutt2021gnina},  and four learning based methods DeepDTA~\cite{ozturk2018deepdta}, Planet~\cite{zhang2023planet}, Banana~\cite{brocidiacono2022bigbind}, and DrugCLIP~\cite{gao2024drugclip}.
\paragraph{Implementation Details}
\label{implementation_details}
In our implementation, we set the hyperparameter $\lambda$ to 0.2. The temperature coefficient $\tau$ is set to 0.07. Each time, we sample 48 protein-molecule pairs for contrastive learning. The code length of the output binary hash codes is 128. Our model is trained on NVIDIA RTX A6000 GPUs, and each model is trained up to 200 epochs. The model is trained for five random seeds and we report the average results. The CASF\--2016~\cite{su2018comparative} dataset is used as the validation set to select the best epoch. We utilize gradient accumulation, performing gradient backpropagation every four steps on a single GPU card,  which is equivalent to using four cards for distributed training. The time test is running on the Intel Xeon Gold 6240R CPUs.
\begin{table}[t]
\caption{Results on DUD-E}
  \label{table1}
  \centering
  \begin{tabular}{l|c|c|ccc}
    \toprule        
                 & AUROC    & BEDROC   & $\mathrm{EF}^{0.5\%}$   & $\mathrm{EF}^{1\%}$   & $\mathrm{EF}^{5\%}$   \\
    \midrule 
    Glide\--SP     & 76.70    & 40.70    & 19.39    & 16.18    & 7.23 \\
    Vina         & 71.60    & \--      & 9.13     & 7.32     & 4.44 \\
    \midrule
    NN\--score     & 68.30    & 12.20    & 4.16     & 4.02     & 3.12 \\
    RF\--Score      & 65.21    & 12.41    & 4.90     & 4.52     & 2.98 \\
    Pafnucy      & 63.11    & 16.50    & 4.24     & 3.86     & 3.76 \\
    Planet       & 71.60    & \--      & 10.23    & 8.83     & 5.40 \\
    Banana       & 50.14    & 2.40     & 1.19     & 1.18     & 1.01 \\
    DrugCLIP     & 79.45    & 47.82    & 37.86    & 30.76    & 10.10 \\
    \midrule
    DrugHash         & \textbf{83.73}    & \textbf{57.16}    & \textbf{43.03}    & \textbf{37.18}    & \textbf{12.07} \\    
    \bottomrule
  \end{tabular}
\end{table}
\subsection{Results}
\paragraph{Evaluation on DUD-E}
We compare the AUROC, BEDORC, and EF of DrugHash with baselines on the DUD\--E dataset. The results are shown in Table~\ref{table1}. DrugCLIP has several versions in the original paper. For a fair comparison, we train DrugCLIP using the same dataset and evaluate the zero-shot results on DUD\--E, and the results are consistent with that of the original paper. DrugHash does also not perform any finetuning on the DUD\--E dataset and evaluates the zero-shot results.  We can find that DrugHash outperforms the baselines across all metrics. Notably, DrugHash has used the same encoder and training data as DrugCLIP, but it outperforms DrugCLIP by a large margin across all metrics, which shows the superiority of our hashing strategy.
\paragraph{Evaluation on LIT\--PCBA}
We compare the AUROC, BEDORC, and EF of DrugHash with baselines on the LIT\--PCBA dataset. The results are shown in Table~\ref{table2}. LIT\--PCBA is a more challenging dataset compared to DUD\--E, and we can find that the performance of all methods has declined. DrugHash outperforms all other methods on metrics of BEDROC, $\mathrm{EF}^{05\%}$ and $\mathrm{EF}^{1\%}$. The performance of DrugHash on AUROC is not outstanding. However, as pointed out in~\cite{truchon2007evaluating}, AUROC is not sensitive to early recognition, which means a successful VS method should rank actives very early among all molecules, because only a small proportion of potential actives will be tested in experiments. So AUROC is less meaningful compared with the other metrics. Although DrugHash may not surpass Banana on $\mathrm{EF}^{5\%}$, it significantly outperforms Banana on $\mathrm{EF}^{05\%}$ and $\mathrm{EF}^{1\%}$, and $\mathrm{EF}^{05\%}$ and $\mathrm{EF}^{1\%}$ are more aligned with the requirement of early recognition than $\mathrm{EF}^{5\%}$. Moreover, Banana generalizes poorly on the DUD-E dataset. Overall, the experiment results show the effectiveness of DrugHash.

\begin{table}[t]
\caption{Results on LIT\--PCBA}
  \label{table2}
  \centering
  \begin{tabular}{l|c|c|ccc}
    \toprule        
               &AUROC   & BEDROC  & $\mathrm{EF}^{0.5\%}$   & $\mathrm{EF}^{1\%}$   & $\mathrm{EF}^{5\%}$   \\
    \midrule 
    Surflex    & 51.47  & \--     & \--    & 2.50    & \--   \\
    Glide\--SP   & 53.15  & 4.00    & 3.17   & 3.41    & 2.01  \\
    Gnina      & 60.93  & 5.40&\--  & 4.63    & \--   \\
    \midrule
    DeepDTA    & 56.27  & 2.53    & \--    & 1.47    & \--   \\
     Planet     & 57.31  & \--     & 4.64   & 3.87    & 2.43  \\  
    Banana     & \textbf{62.78}  & 5.02    & 3.98   & 3.79    & 
    \textbf{2.83}   \\
    DrugCLIP   & 56.36  & 6.78    & 7.77   & 5.66    & 2.32  \\
    \midrule
    DrugHash       & 54.58  & \textbf{7.14}    & \textbf{9.65}   & \textbf{6.14}    & 2.42 \\    
    \bottomrule
  \end{tabular}
\end{table}
\begin{table}[t]
\caption{Comparison of memory and time cost }
  \label{table3}
  \centering
  \begin{tabular}{l|rr|rr}
    \hline
    \multirow{2}{*}{Model}&\multicolumn{2}{c|}{Memory Cost}&\multicolumn{2}{c}{Time Cost}\\    
    & ZINC($\sim$2.3M) & REAL($\sim$6.5B) & ZINC($\sim$2.3M) & REAL($\sim$6.5B)\\
    \hline
    Vina    &\--    &\--    &1863 days    & \--\\
    Planet  &2.57GB    &7264GB    &1749 seconds    & 1408 hours\\
    Banana  &1.00GB    &3080GB    & 67.05 seconds    & 52 hours \\
    DrugCLIP&1.00GB    &3080GB &0.35 seconds    & 1064 seconds\\
    \midrule
    DrugHash&0.03GB   &96GB   &0.20 seconds    & 301 seconds\\
    \bottomrule
  \end{tabular}
\end{table}

\paragraph{Memory and Time cost}
We show the memory and time cost of Planet, Banana, DrugCLIP, and DrugHash when adopting ZINC and Enamine REAL as the target molecule library in Table~\ref{table3}. Unlike the learning-based methods, Vina could only store the raw molecule files. The memory cost of raw molecule files is extremely high, so we do not present them in the table. For the ZINC database, which contains around 2.3 million ready-to-dock molecule files, the memory cost of real-valued vectors is still acceptable for the three real-valued methods. But when the database size increases to the size of the REAL database, which contains 6.5 billion molecules, the memory cost for real-valued vectors rises to more than 3080GB, which is unacceptable for the memory of most computers. However, DrugHash only needs 96GB to store the REAL database, which is about 32 times less than Banana and DrugCLIP, and 75 times less than Planet.

We tested the time cost of the above methods when using only one target protein as the query. The time cost of Vina to dock the total ZINC dataset takes around 1863 days, which is already impractical in real-world applications, so we do not show the time cost for the more massive REAL database. Though Planet and Banana could pre-encode the proteins and molecules to vector representations, they have to further fuse the protein and molecule representations to predict the final binding affinity or interaction. On a database with the same size of ZINC, their required time is still acceptable, but on REAL, their time cost rises to tens or even thousands of hours. DrugCLIP takes much less time on both databases compare with the above methods. DrugHash is the fastest method compared with others and achieves more than three times faster than DrugCLIP on the REAL database. DrugHash has more advantages as the number of the target protein increases.

We show the results of a comprehensive comparison of $\mathrm{EF}^{05}$, memory cost and time cost in Figure~\ref{cmp}. In conclusion, Vina is not practical due to the large memory and time costs. Planet and Banana requires less memory and time cost, but it is still too large for real applications, and their accuracy is not satisfactory. Although DrugCLIP performs VS at a relative faster speed, its pre-encoded molecular vectors result still poses storage challenges for computational devices. DrugHash addresses the issue of high memory consumption required by the other learning-based methods and simultaneously enhances retrieval speed. And DrugHash is the most accurate method compared with the other baselines, which shows the promising potential of DrugHash in VS.

\subsection{Analysis of Hashing Strategy}
\begin{figure}[t]
    \centering
    \subfloat[]{
    \label{cmp}\includegraphics[width=0.52\textwidth]{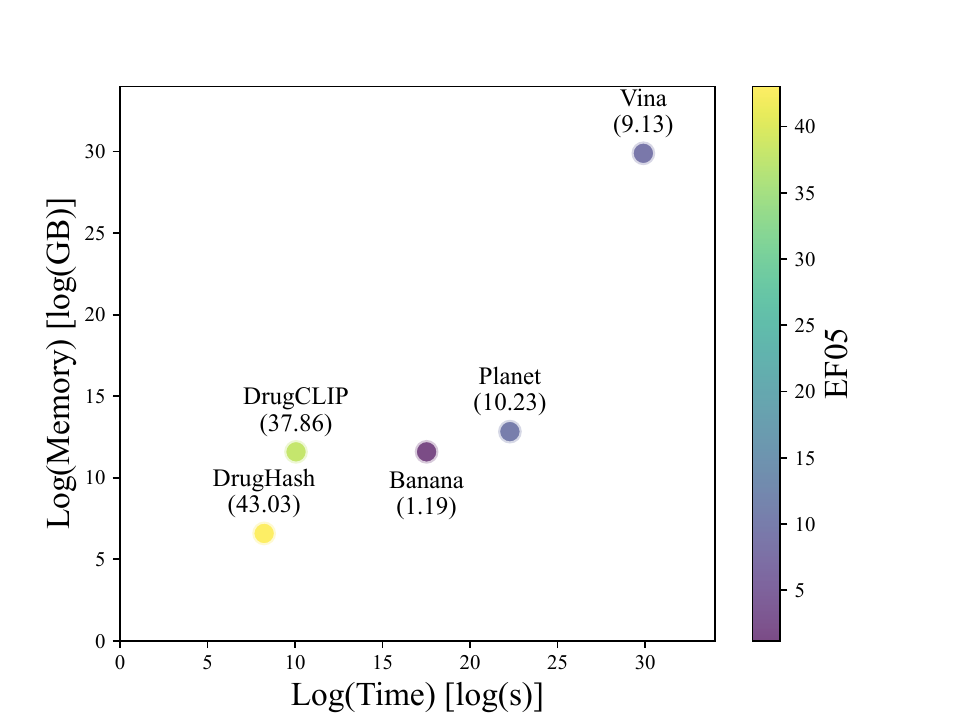}
    }
    \subfloat[]{
    \label{figure3}\includegraphics[width=0.40\textwidth]{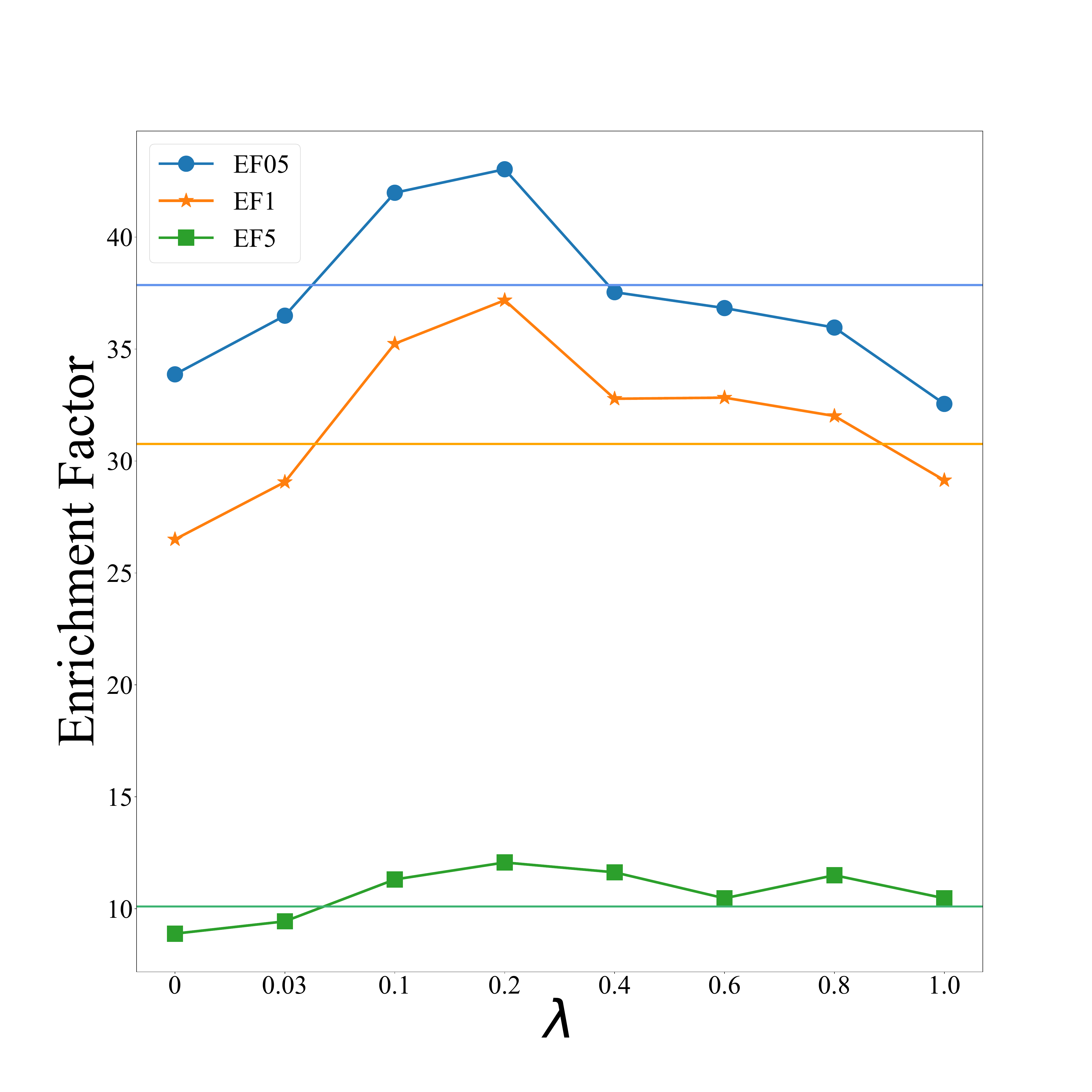}
    }
    \caption{(a) Comprehensive comparison of $\mathrm{EF}^{0.5\%}$, memory cost, and time cost. (b)The results of $\mathrm{EF}^{0.5\%}$,$\mathrm{EF}^{1\%}$ and $\mathrm{EF}^{5\%}$ on different $\lambda$. The horizontal line indicates the results obtained by removing the hashing strategy in training and directly using the real-valued vectors for calculation.}
\end{figure}

In the experiments mentioned above, we demonstrated the comprehensive advantages of our method in terms of accuracy, storage, and retrieval speed for VS. All these improvements can be attributed to the hashing strategy we designed. In this subsection, we will conduct a detailed analysis of how the hashing strategy works.
\paragraph{Sensitivity Analysis of $\lambda$}
We study the sensitivity of the important hyperparameter $\lambda$ in DrugHash. The results are shown in Figure~\ref{figure3}. When we set $\lambda$ to 0, it means we do not adopt any hashing strategy to train the model, but in the evaluation phase, we still use the $\mathrm{sign}(\cdot)$ function to obtain the binary hash codes. It could be seen as the ablation study of the hashing strategy. In this case, we can find that the accuracy dropped by a large margin, which proves the necessity of our hashing strategy. When we tune the $\lambda$ from 0 to 1.0, the accuracy of the DrugHash initially improves and then declines, with the turning point at 0.2.  The sensitivity experiment of $\lambda$ suggests a careful choice of $\lambda$ to make the output of the model closer to binary hash codes while avoiding excessive regularization constraints. When the value of $\lambda$ is between 0.1 and 0.4, the accuracy of binary retrieval can consistently exceed that of real-valued vectors.

\begin{table}[t]
\caption{Results of DrugHash with different code length}
  \label{table4}
  \centering
  \begin{tabular}{l|c|c|ccc}
    \toprule        
      Model   &AUROC    &BEDROC    & $\mathrm{EF}^{0.5\%}$   & $\mathrm{EF}^{1\%}$   & $\mathrm{EF}^{5\%}$   \\
    \midrule
    DrugHash-64&83.17&55.12&41.01&35.40&11.73\\
    DrugHash-128 $\star$    & 83.73    & 57.16    & 43.03    & 37.18    & 12.07 \\ 
    DrugHash-256    &84.96&59.23&43.67&37.76&12.50\\
    DrugHash-512    &84.06&60.50&44.44&39.30&12.57\\
    \bottomrule
  \end{tabular}
\end{table}
\paragraph{Code Length Experiment}
We study the influence of the length of the output binary hash codes. The results are shown in Table~\ref{table4}. DrugHash-128 is marked with $\star$ to denote our original implementation. We study the output code length of $\{64 ,128,256,512\}$, and the accuracy of DrugHash continuously improves as the code length increases. Therefore, the accuracy of DrugHash could be improved by simply increasing code length. However, a larger code length will bring larger memory and time costs. But due to the efficiency of binary hash codes, DrugHash is capable of adopting larger code lengths. Even with the code length of 512, DrugHash requires less than 200GB of memory cost to store the REAL database. DrugHash offers a good trade-off between accuracy and cost in real-world applications.

\begin{figure}[t]
    \centering
    \subfloat[]{
    \label{figure4}\includegraphics[width=0.49\textwidth]{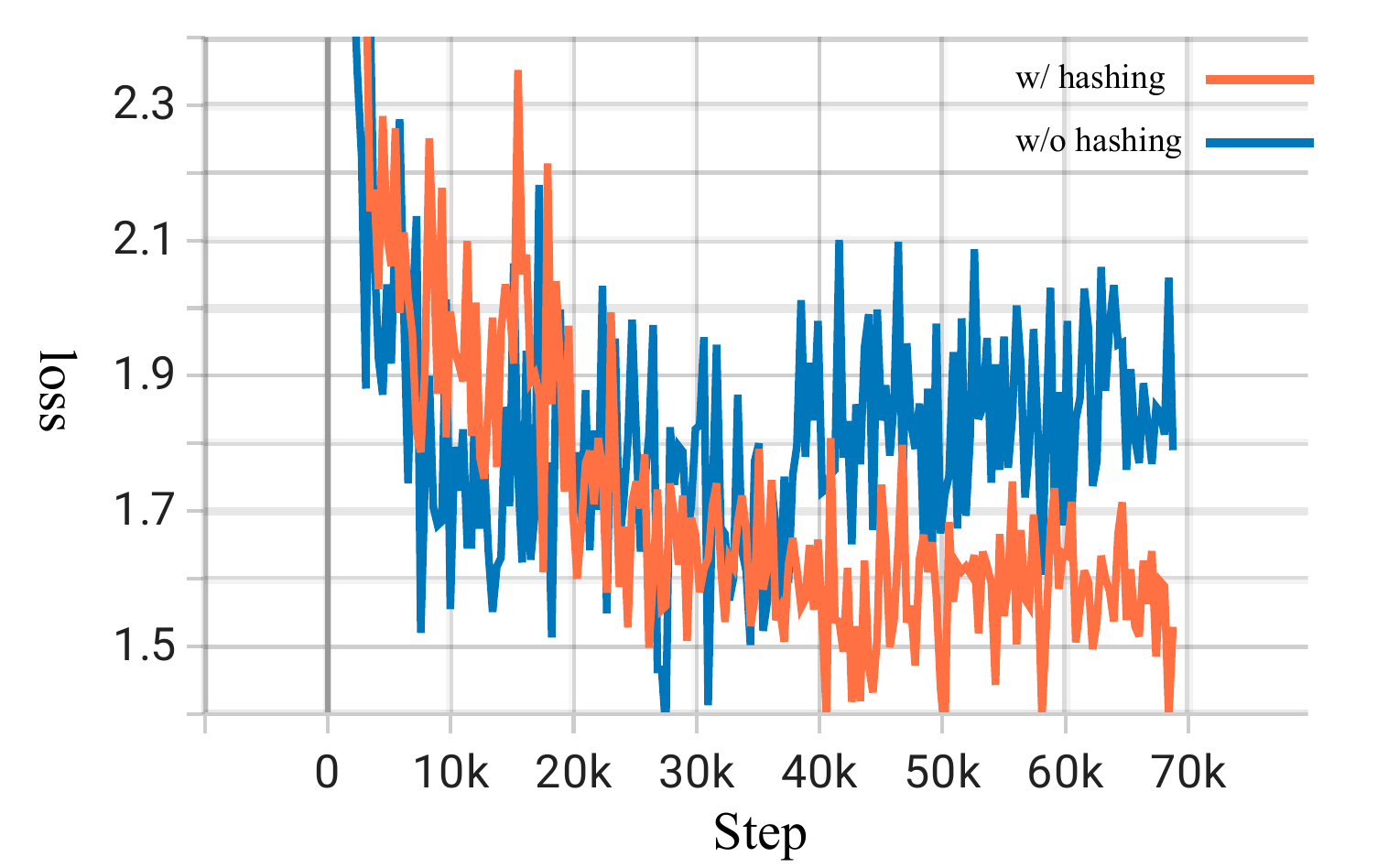}
    }
    \subfloat[]{
    \label{figure5}\includegraphics[width=0.49\textwidth]{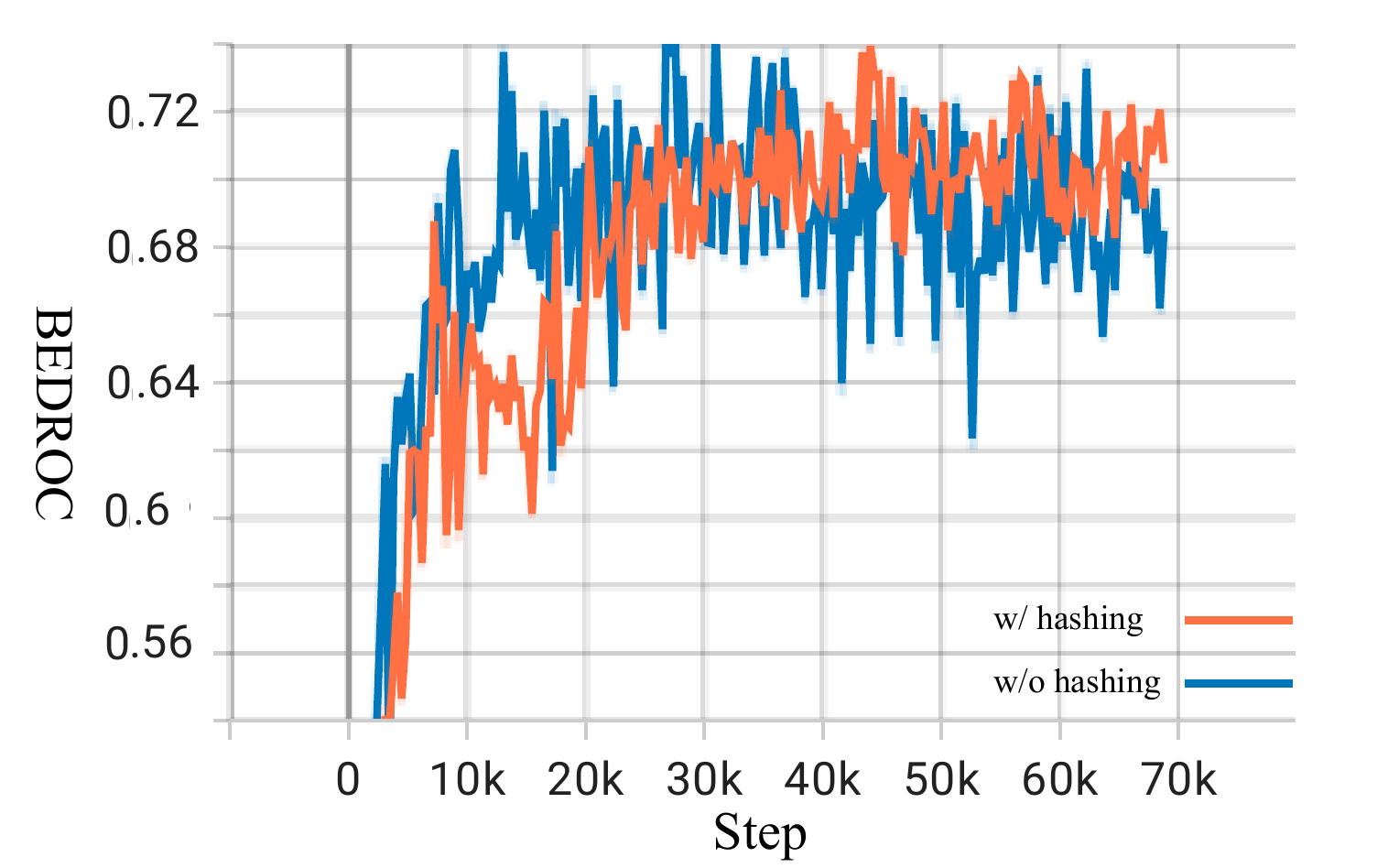}
    }
    \caption{Overfitting analysis of DrugHash with and without hashing strategy. (a) The loss curve on validation set. (b)The BEDROC curve on validation set.}
\end{figure}

\paragraph{Overfitting Analysis}
We provide an explanation of how the hashing strategy enhances model accuracy. We demonstrate the loss and BEDROC curves of DrugHash on the validation set with and without the hashing strategy in Figure~\ref{figure4} and Figure~\ref{figure5}. In Figrue~\ref{figure4}, We can find that the model without the hashing strategy begins to overfit after 35k steps, with the loss on the validation set continuing to increase. However, after adding the hashing strategy, the loss on our validation set continues to decrease and reaches a lower validation loss compared to when hashing is not used. A similar phenomenon can be observed in Figure~\ref{figure5}. The model without the hashing strategy shows a decline in BEDROC in the later stages of training, while the model with the hashing strategy maintains a higher BEDROC. The experiment indicates that the hashing strategy could alleviate the model overfitting and improve the accuracy.

\section{Conclusion}
In this paper we propose a hashing-based contrastive learning method, called DrugHash, for VS.
DrugHash treats VS as a retrieval task that uses efficient binary hash codes for retrieval. In particular, DrugHash designs a simple yet effective hashing strategy to enable end-to-end learning of binary hash codes for both protein and molecule modalities, which can dramatically reduce the memory and time costs with higher accuracy.
Experimental results show that DrugHash can outperform other existing methods to achieve state-of-the-art accuracy, with a memory saving of 32$\times$ and a speed improvement of 3.5$\times$. We also conduct detailed experiments about how the hashing strategy works.
\section{Limitations \& Broader Impact}
\label{limitations}
There exists some limitations in DrugHash. Firstly, in DrugHash, we designed a simple yet effective hashing strategy to conduct a preliminary exploration of the efficacy of hashing in VS tasks. Actually, more hashing strategies could be explored to address the binarization problem. Secondly, the effect of the size of training data could be further explored. In this paper, we use PDBBind as the training set. Larger complex datasets like BioLip and ChEMBL could be used to train the model. Thirdly, since different encoders could be adopted in DrugHash, a more expressive encoder could be designed. We leave the above directions for future work.

This paper presents work with the goal to advance the field of Machine Learning. There are no negative potential societal consequences of our work.

\bibliography{ref}

\end{document}